\documentclass[letterpaper, 10pt, conference]{ieeeconf}  

\IEEEoverridecommandlockouts                              

\usepackage[backend=biber,style=ieee,natbib=true,maxcitenames=1,mincitenames=1]{biblatex} 

\AtEveryBibitem{\clearfield{url}}
\AtEveryBibitem{\clearfield{doi}}
\AtEveryBibitem{\clearfield{urldate}}
\DeclareFieldFormat{urldate}{}

\usepackage{graphicx}
\usepackage{amsmath} 
\usepackage{amssymb}  
\usepackage{algorithm}
\usepackage[noend]{algpseudocode}
\usepackage{hyperref}

\usepackage{balance}
\usepackage{subfig}
\hypersetup{
breaklinks=true,letterpaper=true,colorlinks,linkcolor=[RGB]{153,27,30},citecolor=[RGB]{153,27,30},urlcolor=[RGB]{153,27,30}
    }

\usepackage{amsfonts}
\usepackage{multicol}
\usepackage[normalem]{ulem}
\usepackage{booktabs}

\newcommand{\norm}[1]{\left\lVert#1\right\rVert}
\usepackage[font=small]{caption}

\expandafter\def\expandafter\normalsize\expandafter{%
    \normalsize%
    \setlength\abovedisplayskip{3pt}%
    \setlength\belowdisplayskip{3pt}%
}

\usepackage{titlesec}
\titlespacing*{\section}{0pt}{4px}{4px}

\title{\LARGE \bf
AutoFocus-IL: VLM-based Saliency Maps for Data-Efficient\\ Visual Imitation Learning without Extra Human Annotations
}

\author{Litian Gong$^{1}$, Fatemeh Bahrani$^{2}$, Yutai Zhou$^{2}$, Amin Banayeeanzade$^{2}$, Jiachen Li$^{1}$, Erdem Bıyık$^{2}$
\thanks{$^{1}$ Department of Electrical and Computer Engineering, University of California, Riverside, USA.}
\thanks{$^{2}$ Thomas Lord Department of Computer Science, University of Southern California, USA.}
\thanks{Corresponding Emails:
\tt{lgong024@ucr.edu}, \tt{biyik@usc.edu}.}}

\begin{document}

\maketitle
\thispagestyle{empty}
\pagestyle{empty}

\begin{abstract}

    We present AutoFocus-IL, a simple yet effective method to improve data efficiency and generalization in visual imitation learning by guiding policies to attend to task-relevant features rather than distractors and spurious correlations. Saliency regularization has emerged as a promising way to achieve this, but existing approaches typically require costly supervision such as human gaze data or manual saliency annotations. In contrast, AutoFocus-IL leverages vision-language models (VLMs) to automatically identify and track key objects in demonstrations, generating temporal saliency maps that highlight causal visual signals while suppressing distractors. These maps are then used to regularize behavior cloning policies, yielding stronger alignment between visual attention and task-relevant cues. Our findings highlight that VLM-driven saliency provides a scalable, annotation-free path toward robust imitation learning in robotics. Particularly, our experiments in both the CARLA simulator and real-robot manipulation tasks demonstrate that AutoFocus-IL not only outperforms standard behavior cloning but also surpasses state-of-the-art baselines that assume privileged access to human supervision, such as gaze data. The supplementary materials, including code, datasets, and trained policy videos, are publicly available at \url{https://AutoFocus-IL.github.io/}.

\end{abstract}

\section{Introduction}
\label{sec:introduction}

Imitation learning (IL) has emerged as a central paradigm in robotics, enabling agents to acquire complex behaviors from expert demonstrations \citep{osa2018algorithmic, argall2009survey}. Recent advances leverage large vision-language-action (VLA) models \citep{zitkovich2023rt2,black2024$p_0$,shukor2025smolvla}, trained through behavior cloning (BC) on robot demonstration datasets, to bridge perception and action. While promising, this approach remains bottlenecked by the limitations of robot data collection \citep{saxena2024what, mandlekar2021what}, leading to several fundamental challenges that hinder scalability and real-world deployment.

A primary issue is \emph{data efficiency}. Unlike images or language \citep{deng2009imagenet, common}, robotic data is expensive and slow to collect, requiring physical interaction with the environment and human teleoperation. Estimates suggest that reaching the same scale as today’s language or vision corpora would take over a century of robot operation \citep{goldberg2025good}. Consequently, designing imitation learning methods that make better use of limited data is essential.

This data scarcity further compounds the challenge of \emph{generalization} \citep{gao2025taxonomy}. Existing VLA models consistently show large performance gains when fine-tuned on in-domain data \citep{kim2025finetuning}, revealing their reliance on task- and environment-specific supervision. In practice, this means an end-user may need to provide hundreds of demonstrations for a new task, severely limiting real-world applicability. Since both pretraining and fine-tuning typically rely on behavior cloning \citep{pomerleau1988alvinn}, any improvements in BC would directly enhance policy generalization.

Another persistent issue in imitation learning is \emph{causal confusion} \citep{deHaan2019causal}, where models make predictions based on spurious correlations between visual cues and expert actions. For example, they may attend to background distractors that co-occur with certain actions but are irrelevant to the task \citep{park2021object}. This not only degrades policy robustness but also makes policies brittle in unseen environments.

\begin{figure}
    \centering
    \includegraphics[width=0.9\linewidth]{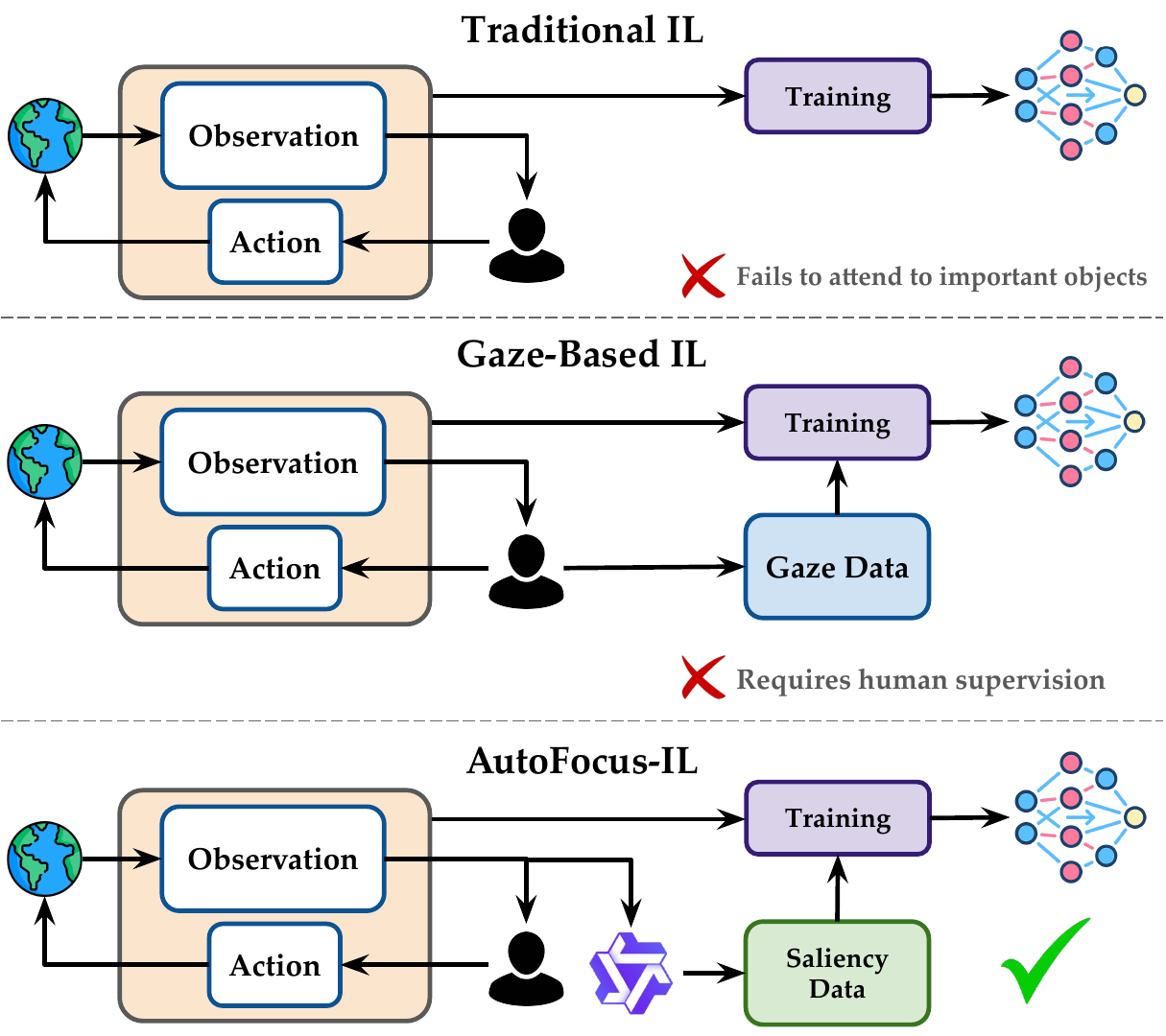}
    \vspace{+1mm}
    \caption{An overview of three different approaches: While traditional imitation learning suffers from causal confusion, gaze-based IL solves this by utilizing an expensive solution of collecting human eye gaze data. However, {AutoFocus-IL} resolves this issue by getting a saliency map annotated by a VLM to retain the benefits of gaze-based IL without incurring the extra data collection costs.}
    \label{fig:intro_fig}
    \vspace{-4mm}
\end{figure}

In this work, we propose a simple yet effective method, AutoFocus-IL (\textbf{Auto}matically \textbf{F}ocusing \textbf{o}n \textbf{c}ontext-relevant objects \textbf{u}sing VLM \textbf{s}upervision for \textbf{I}mitation \textbf{L}earning), to preprocess demonstrations and regularize imitation learning without requiring any additional human annotations. Our approach improves data efficiency, enhances generalization, and mitigates causal confusion by explicitly highlighting task-relevant visual signals for the learner (see \autoref{fig:intro_fig}).

Concretely, we leverage open-source vision-language models to identify objects that are important for the demonstrated task. These candidate key objects are then tracked across the trajectory while also allowing for disappearances or newly introduced objects. Using these detections, we generate saliency masks that emphasize the key objects while masking distractors. The resulting masks are then used to regularize the attention of the policy network during behavior cloning. Experiments on autonomous driving in the CARLA simulator \citep{dosovitskiy2017carla} show \textbf{104\%} improvement over standard BC, while real-world experiments with a WidowX robot arm show \textbf{50\%} improvement. Notably, our method not only surpasses BC but also outperforms baselines that assume access to privileged human annotations, such as gaze data, to identify the relevant and salient parts of the scene.

Overall, our contributions are threefold:
\begin{enumerate}
    \item We introduce a context-aware VLM-driven object filtering pipeline for identifying task-relevant visual signals without additional supervision.
    \item We propose a temporal saliency modeling approach that translates object tracks into structured signals.
    \item We demonstrate that regularizing imitation learning with saliency maps produced with our approach substantially improves data efficiency, generalization, and robustness in both simulation and real-world robot experiments.
\end{enumerate}

\section{Related Work}
\label{sec:related_work}

\subsection{Data-Efficiency in Imitation Learning}


A common approach to address data efficiency is to employ active learning or interactive IL frameworks \citep{celemin2022interactive, settles2009active}, where the agent queries a human expert in order to resolve uncertainties or increase state coverage during training \citep{ross2011reduction, kelly2019hgdagger}.
These strategies improve sample efficiency by reducing redundant demonstrations, but they require a human to remain in the loop throughout training, limiting scalability. In contrast, our method improves data efficiency without any additional expert queries: we preprocess existing demonstration datasets to automatically highlight task-relevant regions, effectively augmenting the supervision signal at no extra human cost.

Methods that target generalization across tasks and domains often incorporate data augmentation \citep{chen2023genaug, laskin2020reinforcement}, domain randomization \citep{tobin2017domain}, or auxiliary tasks for visual representation learning \citep{laskin2020curl}. While effective to some extent, these methods typically remain agnostic to which parts of the scene are semantically important for the task. By explicitly filtering and emphasizing key objects in demonstrations, our approach complements these generalization techniques by ensuring that the model allocates capacity to causal features rather than incidental correlations.

Another line of work attempts to have the model focus only on relevant features to tackle causal confusion—the tendency of policies to misattribute spurious correlations as the root cause of expert behavior \cite{deHaan2019causal}. This issue not only degrades robustness under distribution shift but also reduces data efficiency and model interpretability. To mitigate causal confusion, prior works have focused on expert interventions \cite{deHaan2019causal}, object-centric feature dropout \cite{park2021object}, or keyframes \citep{wen2021keyframefocused}. These methods force the learner to discard distractors and rely on causal signals. However, they often require access to additional expert supervision, or they assume simplified environments. Our method shares the spirit of causal deconfounding, but achieves it in a scalable way by using VLM-guided filtering to identify and focus only on task-relevant objects, without relying on privileged annotations or interventions.

In summary, prior work has made progress on improving data efficiency, generalization, and mitigating causal confusion in IL, but often at the cost of additional supervision or restrictive assumptions. Our method instead achieves these goals through an automatic preprocessing pipeline, requiring no extra human effort and operating directly on raw high-dimensional demonstration data in real-world settings.

\subsection{Vision-Language Model Supervision for Robot Control}

VLMs trained on Internet-scale data provide rich semantic and spatial knowledge, making them a powerful yet scalable alternative to human supervision for robotics.
One particular line of research uses VLMs to generate weak labels for demonstrations. For example, NILS \cite{blank2024scaling} leverages VLMs to automatically segment long-horizon trajectories, detect object-centric events, and emit instruction labels at scale, improving policy learning without human annotation. 

Another prominent direction is reward modeling. VLMs have been applied as zero-shot success detectors \citep{du2023vision, rocamonde2024vision}, providing episodic feedback for reinforcement learning without privileged state information. Others have used VLMs to estimate task progress \cite{sontakke2023roboclip, ma2023vip, zhang2025rewind}, generate language-conditioned rewards \cite{ma2023eureka, yu2023language}, or elicit preference feedback to train reward functions \cite{wang2024rl}.
Beyond labeling and rewards, VLMs have also been employed for high-level task reasoning, grounding, and planning \citep{huang2023voxposer, ling2025impact, ahn2022can, driess2023palme}. These methods demonstrate the versatility of VLM supervision in shaping robot behavior, but they often remain detached from the imitation learning pipeline itself.

Rather than using VLMs to produce episodic labels or high-level plans, our work leverages them to provide context-aware supervision on individual input frames. This enables us to provide dense supervision to regularize the learner’s visual attention toward causal signals, leading to improved data efficiency and generalization without requiring privileged information or additional human effort.

\subsection{Human Supervision for Generating Saliency Maps}
Saliency map has been a widely used tool for post-hoc interpretation of neural networks, and has also been used as a training supervision for enhancing model explainability \citep{zeiler2014visualizing, ismail2021improving}.
A prominent line of work enhances imitation learning by explicitly incorporating human-provided saliency supervision. For example, ViSaRL \cite{liang2024visarl} introduces a framework where annotators mark task-relevant regions in demonstrations, which are then used to learn better state representation for policy learning. Such explicit annotations help encourage models to attend to causal features rather than spurious correlations. However, collecting large-scale saliency labels is costly and prone to human error.

Given the expense of manual annotations, many works instead turn to human gaze as a natural and implicit source of saliency (see \autoref{fig:intro_fig}). Human gaze is well known to correlate with task-relevant features \cite{tanner2019top, admoni2017social, zhang2020human} and can be captured during task execution without additional labeling effort. In robotic manipulation, \citet{kim2020using, kim2022memorybased} showed that gaze supervision enables policies to ignore irrelevant distractors and focus on task-relevant objects.
More generally, gaze-derived attention maps have been leveraged to regularize policies across domains, including robotics, driving, and video games \citep{chen2019gaze,zhang2018agil,saran2021efficiently, biswas2024gaze, banayeeanzade2025gabril}.

Despite their effectiveness, both explicit saliency annotation and gaze supervision face practical limitations. Manual saliency labeling is slow, expensive, and difficult to scale, while gaze data requires specialized hardware and careful experimental design to avoid noise or distraction effects \citep{liu2022eye, chennamma2013survey}. Besides, incorporating gaze as a model of human attention requires very careful tuning because gaze is often directed at a single point in the scene, and humans have peripheral vision that is difficult to model \citep{vater2022peripheral, larson2009contributions}. Our method circumvents these challenges by replacing human supervision with automatically generated saliency. Instead of relying on gaze or manual labels, we leverage VLM-driven object filtering and temporal modeling to produce saliency maps directly from raw sensory inputs. This yields many of the benefits of human-derived saliency: better data efficiency, generalization, and reduced causal confusion, all without the need for any additional human annotation.

\section{Methodology}
\label{sec:methodology}

\begin{figure}[t]
  \centering
  \includegraphics[page=1,width=\linewidth]{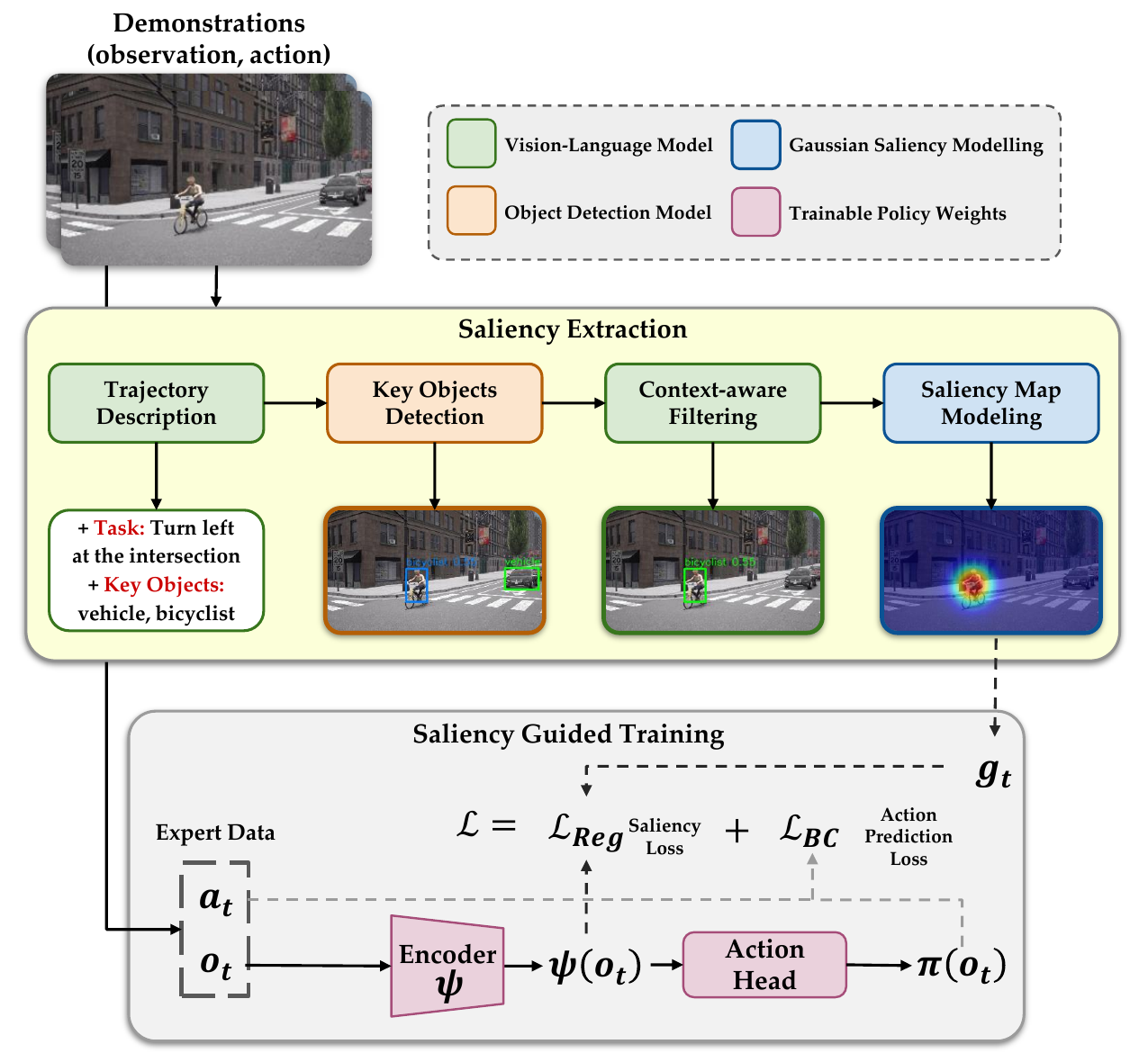}
  \caption{Overview of the AutoFocus-IL pipeline.}
  \vspace{-10px}
  \label{fig:saliency_generation}
\end{figure}

We propose AutoFocus-IL, a pipeline that automatically extracts task-relevant objects from demonstrations and uses them to generate saliency maps for imitation learning. The method consists of three components: (1) context-aware VLM filtering that identifies and tracks key objects across demonstration, (2) temporal saliency modeling that converts object tracks into smooth multi-peak saliency maps, and (3) a regularization technique that leverages saliency for policy learning. \autoref{fig:saliency_generation} provides an overview of our method.

\subsection{Context-Aware VLM Filtering}
Each demonstration is a trajectory of image-action pairs $\xi=(o_0, a_0, o_1, a_1, \dots)$, where $o_t \in [0,1]^{w\times h\times 3}$ is the visual observation at time $t$ and $a_t$ is the expert action. Our first goal is to annotate each observation $o_t$ with a saliency map $g_t \in [0,1]^{w\times h}$ where each pixel denotes how important the corresponding pixel of $o_t$ is for the task. For example, a pixel that belongs to a cloud should have a low saliency value for the autonomous driving task, whereas the pixels of a stop sign should be labeled as highly salient when the car is approaching an intersection.

The straightforward way of automatically getting these labels would be to query the VLM with each and every observation. However, this approach is not easily scalable because processing every single frame with a VLM would incur a prohibitively high computational cost. Instead, we first perform equidistant frame sampling with stride $s$ to select a small subset of frames from each demonstration.

Another challenge of using VLMs to generate saliency maps is due to the uncertainty about the objects and the task. Since we are not making any assumptions about the expert demonstration dataset, we only have a set of trajectories without any language label of what the task is or what objects are present in the scene. This creates an ambiguity: if we are trying to train a robot arm to pick up a carrot, the carrot is obviously relevant; in the same scene, the task might be dicing a tomato, in which case the carrot should not be labeled as salient.

To resolve this, we first query the VLM with the entire subset of frames that were sampled from the full demonstration. In this work, we use Qwen2.5-VL-Instruct-72B \cite{bai2025qwen2_5_vl} as the VLM. We ask it to generate\footnote{Detailed instructions used to prompt the VLM can be found in the supplementary material, available on the \href{https://autofocus-il.github.io/}{project website}.}:
\begin{enumerate}
    \item A \emph{global context summary}, $c$, of the trajectory, including descriptions of the task, the environment, and potential risks.
    \item A \emph{candidate object vocabulary} $\mathcal{V}$, i.e., the set of objects that could appear in the demonstration.
\end{enumerate}
While the latter is useful for detecting and tracking the objects in the scene, the former will be used to determine their relevance to the task.

\textbf{Detection and Tracking.} Using the object vocabulary $\mathcal{V}$, we apply an open-set object detector (in this work, Grounding DINO \cite{liu2024grounding}) to every frame in the trajectory. The detector outputs bounding boxes and confidence scores for all recognized objects. To maintain consistent object identities across time, we perform frame-to-frame association via the Hungarian method \citep{kuhn1955hungarian} based on bounding box overlap (Intersection-over-Union) in consecutive frames\footnote{We chose to use Grounding DINO with the Hungarian method for tracking the detected objects due to its computational efficiency. Alternative implementations could use object tracking models from computer vision (e.g., \cite{cheng2023tracking,ren2024grounded}) or other traditional tracking algorithms from the image processing literature (e.g., \cite{baykara2017real,sengar2017moving}).}. This yields continuous object tracks, each with a persistent ID.


Because the set of visible objects can change during a demonstration (objects enter or leave the scene), we divide the trajectory into contiguous sub-sequences such that within each sub-sequence, the set of tracked object IDs remains constant. This segmentation of the trajectory allows the VLM to reason about objects in a locally stable context.

\textbf{Filtering with VLM supervision.} After obtaining the list of task-relevant objects in the scene, we filter them based on their importance. For example, a pedestrian is always task-relevant for autonomous driving, but should only affect autonomous driving decisions if they are on the road; in contrast, a pedestrian walking on the sidewalk is not critical and should not affect the driving system.
We achieve this filtering with the VLM: for the first frame of each sub-sequence, we construct a context prompt that consists of the frame image $o_t$, the expert action $a_t$, the global context summary $c$, and the list of active objects with their IDs and bounding boxes.
The VLM is then asked to select the \emph{key objects} that are important for deciding to take action $a_t$ at that timestep.

Objects not marked as key by the VLM are pruned for the entire sub-sequence, ensuring consistency across time.
If the VLM detects that important objects are missing (e.g., due to detection failure), it flags a set of missing categories. The detector is then re-run with the augmented vocabulary until either the missing objects are recovered or a small iteration cap is reached.
At the end of this stage, each frame is annotated with the set of task-relevant object bounding boxes. These serve as the basis for downstream saliency modeling.

\subsection{Temporal Multi-Peak Saliency Modeling}

The filtered object tracks indicate which objects matter, but not how strongly the learner should attend to them across time. We therefore convert the key-object annotations into continuous saliency maps. Namely, for each timestep $t$, we collect the centers of all key objects detected in the observations $o_t, o_{t-1}, \dots, o_{t-t'}$ where $t'$ is a hyperparameter. To construct the saliency map $g_t$, we add a Gaussian distribution around each object center. Temporal weighting is applied so that more recently detected objects are sharper and more influential, while older detections are more blurred and less influential.

Mathematically, we perform the following operation to obtain the saliency map at any pixel location $p$:
\begin{align}
    \bar{g}_t(p) = \sum_{k=0}^{t'} \alpha^k \sum_{i} \exp\!\Big( -\tfrac{\|p-\mu^{(i)}_{t-k}\|_2^2}{2\gamma^2\beta^{-2k}} \Big), 
    \label{eq:pre_norm_past}
\end{align}
where $\alpha,\beta\in[0,1]$, $\gamma>0$ are hyperparameters, and the inner summation is over the key objects in the sub-sequence whose center locations are denoted by $\mu$. The variance increases from the current frame $t$ to the past, producing broader kernels for older frames. We also weight the Gaussian kernel at each timestep by a temporal decay factor $\alpha$. 

Finally, normalizing $\bar{g}_t$ over all pixels into the range $[0,1]$ yields a dense saliency map $g_t$ for frame $t$.
This multi-peak temporal model preserves evidence from multiple objects across time while naturally fading older observations, thereby mimicking human-like sustained attention. Also, by retaining information from previous frames, we enhance the quality of saliency maps by mitigating the impact of missed detections in the current frame.



\subsection{Saliency-Guided Regularization}
Following the work of \cite{banayeeanzade2025gabril}, we integrate the VLM-generated saliency maps with behavioral cloning via a knowledge distillation approach \citep{hinton2015distilling}. In fact, we employ a convolutional neural network (CNN)-based policy $\pi$ trained to map images $o_t$ to actions $a_t$ via supervised learning, and regularize the intermediate feature maps of the policy using the VLM-generated saliency maps.
Let $\psi(o_t) \in \mathbb{R}^{h \times w \times c}$ be the CNN feature map representation of an observation $o_t$, upsampled to the input resolution $w\times h$, with $c$ channels. We minimize the $\ell_2$-norm of activation map features modulated by the saliency map $g_{t}$, with the goal of penalizing feature activations in regions not covered by the saliency map.
The final training loss is:
\begin{align}
    \mathcal{L} = \mathcal{L}_{BC}(\pi,o_t,a_t) + \lambda\norm{(1-g_t)\odot\psi^2(o_t)}_2,
    \label{eq:loss}
\end{align}
where $\odot$ is the element-wise matrix product. The first term is the standard behavior cloning loss for observation $o_t$ with expert action $a_t$ (e.g., mean squared error for continuous action spaces or cross-entropy for discrete action spaces), and the second term performs the regularization with the VLM-generated saliency map, with its strength scaled by the hyperparameter $\lambda$.
Note that AutoFocus-IL computes saliency only during training; at inference time, no additional saliency predictor is required, in contrast to several prior studies \citep{liang2024visarl, zhang2018agil, chen2019gaze}.

\section{Simulation Experiments}
\label{sec:sim}

\subsection{Environment Setup}
\begin{figure}[t]
    \centering
    \includegraphics[width=.99\linewidth]{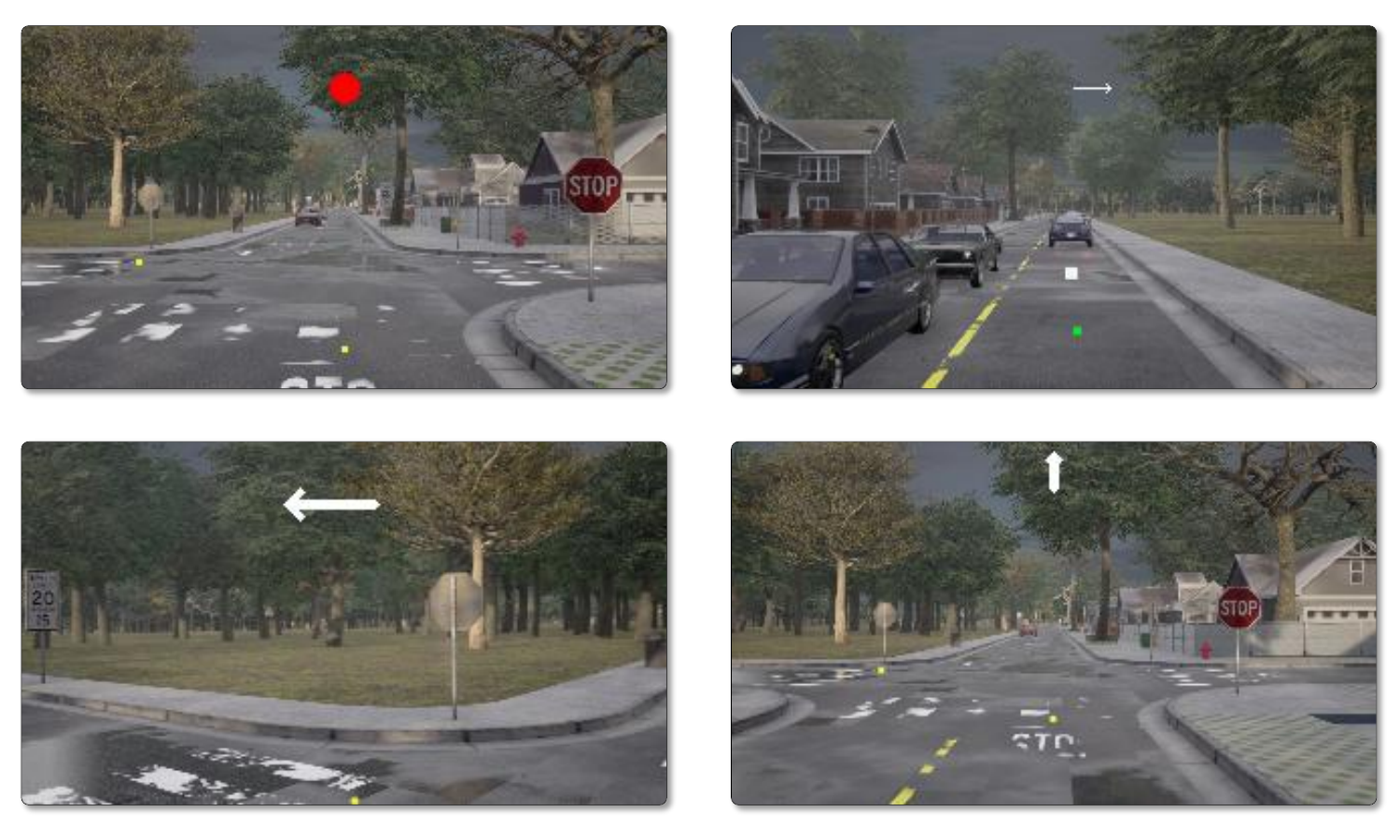}
\vspace{-15px}
\caption{Confounded overlay visualization in the CARLA simulator. Action-conditioned icons are rendered along the top margin of each frame to induce spurious correlations, while leaving the underlying dynamics and expert labels unchanged. The red circle simulates a brake light, while arrows denote the steering direction, with their thickness indicating the throttle applied in the previous timestep.}
    \vspace{-15px}
    \label{fig:confound_vis}
\end{figure}

We evaluate AutoFocus-IL in the CARLA urban driving simulator \cite{dosovitskiy2017carla} using the Bench2Drive benchmark \cite{jia2024bench2drive}. 
We consider two observation settings: 
(i) \emph{Original}, where the policy receives raw RGB images, and 
(ii) \emph{Confounded}, where we introduce action-conditioned icons on top of each frame. 
These overlays do not affect dynamics or expert actions, but introduce spurious correlations intended to test robustness against causal confusion (\autoref{fig:confound_vis}). This is a modification strategy adopted by several prior studies, e.g., \cite{banayeeanzade2025gabril,park2021object}.

We use the dataset provided by \citet{banayeeanzade2025gabril} to show AutoFocus-IL can improve imitation learning on datasets that are readily available on the Internet. The training data consists of 10 routes with 20 expert trajectories per route. 
Evaluation is conducted on two disjoint sets: 
(1) \emph{seen} routes, which are the same 10 routes used for training but with 2 different random seeds (20 evaluations per method), and 
(2) \emph{unseen} routes, a disjoint set of 10 held-out routes, also with 2 seeds each (20 evaluations per method).

Input images are resized to $320\times 180\times 3$ (width $\times$ height $\times$ channels). All baselines share the same backbone CNN encoder and optimizer configuration; auxiliary heads or additional loss terms are only included if explicitly required by the baseline. 
This ensures differences in performance arise from the saliency supervision rather than from architecture or optimization choices.

\subsection{AutoFocus-IL Hyperparameters}

For each demonstration, we query the VLM on a fixed set of 25 uniformly sampled frames for global context and vocabulary generation, i.e., $s=25$. For the Gaussian distributions in saliency map generation (\autoref{eq:pre_norm_past}), we let $\alpha=0.7$, $\beta=0.8$, $\gamma=15$, and $t'=4$.

We use the final convolutional feature map of the policy network as $\psi$ in \autoref{eq:loss} and let $\lambda=10$. Both behavior cloning loss and saliency regularization are applied to all frames, except in an ablation study, where the saliency regularization is applied to the prescribed fraction of frames.

\subsection{Baselines and Metric}

We compare AutoFocus-IL against
plain {BC} with no saliency regularization. In addition, we compare it against baselines that use human gaze as a regularizer, thanks to the availability of gaze information in the dataset, which was not used for AutoFocus-IL. These baselines are the following: 
{GMD} \cite{chen2019gaze}, {ViSaRL} \cite{liang2024visarl}, {GRIL} \cite{thakur2021gril}, {AGIL} \cite{zhang2018agil}, {GABRIL} \cite{banayeeanzade2025gabril}, and {GABRIL+GMD} which was implemented by \citet{banayeeanzade2025gabril} to incorporate {GMD} into their framework. 

Performance is measured by the \emph{driving score} from CARLA Leaderboard 2.0  \cite{leaderboard_v2}, a scalar metric that aggregates success rate, safety, and efficiency. 
Scores are averaged across seeds and routes for each evaluation split.

\begin{figure}[t] 
    \centering
    \subfloat[Original: Seen Routes]{%
        \label{fig:carla_original_seen}\includegraphics[width=0.49\columnwidth]{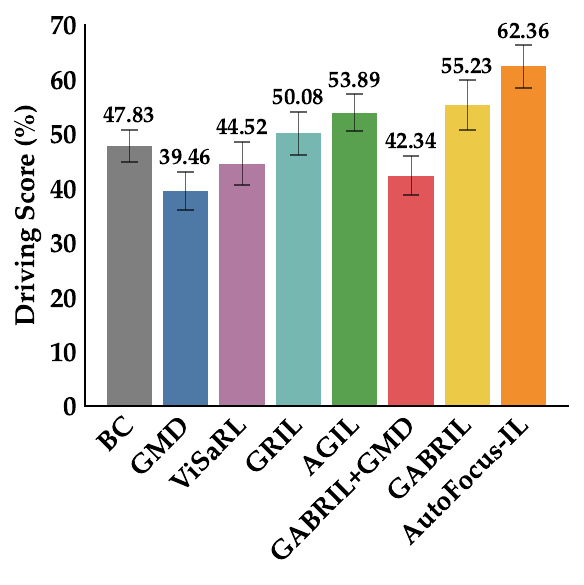}
    }
    \subfloat[Original: Unseen Routes]{%
        \label{fig:carla_original_unseen}        \includegraphics[width=0.49\columnwidth]{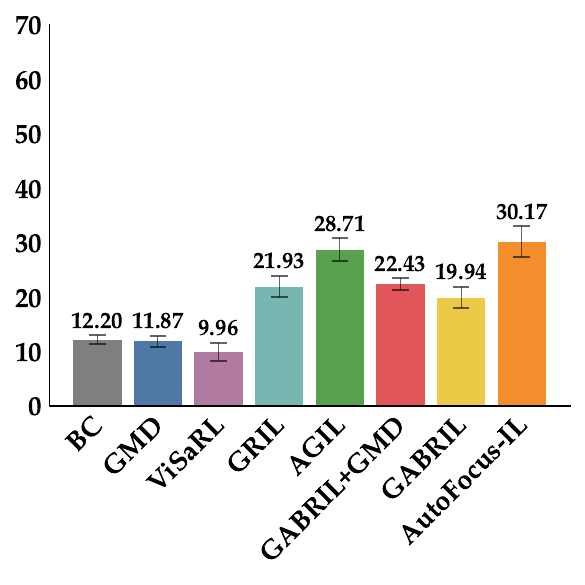}
        }
        
    \subfloat[Confounded: Seen Routes]{%
        \label{fig:carla_conf_seen}        \includegraphics[width=0.49\columnwidth]{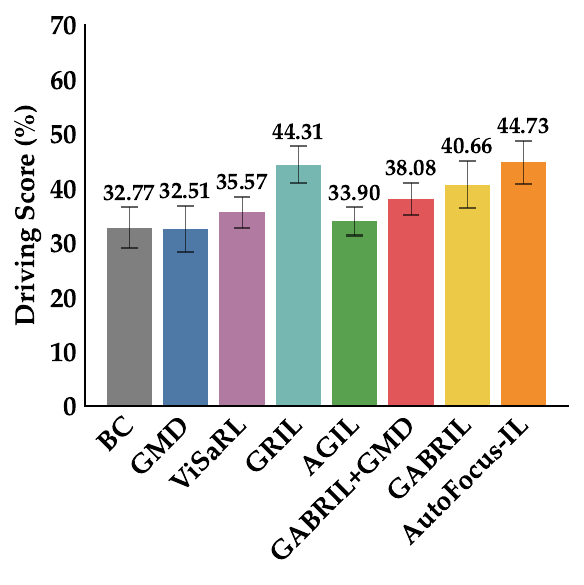}
    }
    \subfloat[Confounded: Unseen Routes]{%
        \label{fig:carla_conf_unseen}
        \includegraphics[width=0.49\columnwidth]{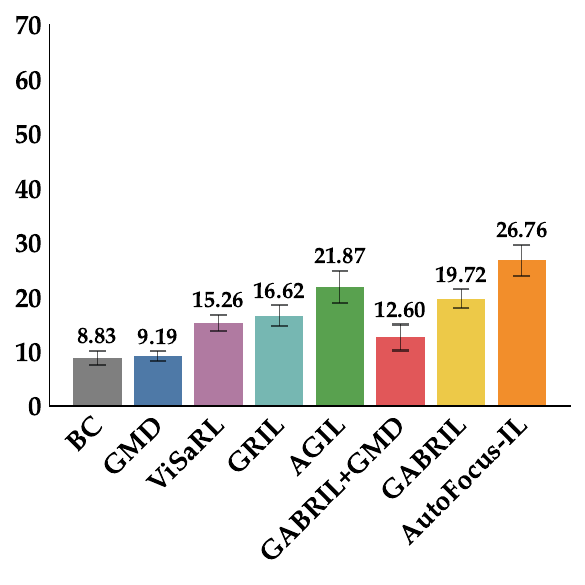}
        }
    \vspace{-3px}
    \caption{CARLA Driving Score (mean $\pm$ standard error) on seen and unseen routes in both original and confounded environments. Except BC, other baselines use human gaze to improve imitation learning, while AutoFocus-IL uses VLM-generated saliency maps.}
    \label{fig:carla_exp}
    \vspace{-5mm}
\end{figure}

\subsection{Main Results}

\autoref{fig:carla_original_seen} and \autoref{fig:carla_original_unseen} show that AutoFocus-IL improves over {BC}. It attains a driving score of \textbf{62.36\%} on seen routes and \textbf{30.17\%} on unseen routes, whereas these numbers are only \textbf{47.83\%} and \textbf{12.20\%} for {BC}. This shows that AutoFocus-IL is an effective method that processes existing imitation learning datasets to achieve better results with the easy-to-implement saliency regularization.

Even more impressively, AutoFocus-IL outperforms baselines that have access to human gaze data for resolving causal confusion or improving data efficiency. Without access to this privileged information, AutoFocus-IL improves driving scores by more than $\mathbf{2.2}$ points over the top baseline in seen routes, and by more than $\mathbf{1.7}$ points in unseen routes.

In \autoref{fig:carla_conf_seen} and \autoref{fig:carla_conf_unseen}, we present the results from the confounded settings with action-conditioned overlays. As expected, this setting leads to lower driving scores because of causal confusion. However, AutoFocus-IL remains the best performing method in both seen and unseen routes. 
While {GRIL} achieves competitive performance in seen routes, AutoFocus-IL generalizes better and attains a much higher driving score in unseen routes. It is worth emphasizing again that {GRIL} and all other baselines except {BC} utilize human gaze/saliency information, which requires additional human annotations or hardware, whereas AutoFocus-IL's saliency maps are generated automatically with the VLM.


\subsection{Ablation Studies}

\textbf{Temporal vs.\ non-temporal saliency.} We compare our temporal multi-peak saliency modeling against a non-temporal, frame-wise variant. 
As shown in \autoref{tab:temporal}, temporal modeling consistently improves results across all splits, yielding an average gain of $\mathbf{12.7}$ points in the driving score.

\begin{table}[tb]
\centering
\caption{CARLA Driving Score (mean $\pm$ standard error) obtained from using Temporal vs. Non-Temporal Saliency modeling.}
\small
\setlength{\tabcolsep}{6pt}
\renewcommand{\arraystretch}{1.15}
\begin{tabular}{lcc}
\toprule
\textbf{Setting} & \textbf{Non-Temporal} & \textbf{Temporal} \\
\midrule
Seen—Original      & \textbf{$49.83 \pm 3.22$}  & \textbf{$62.36 \pm 3.89$} \\
Seen—Confounded    & \textbf{$34.93 \pm 3.65$}  & \textbf{$44.73 \pm 3.94$} \\
Unseen—Original    & \textbf{$16.04 \pm 1.73$}  & \textbf{$30.17 \pm 2.83$} \\
Unseen—Confounded  & \textbf{$12.55 \pm 0.88$}  & \textbf{$26.76 \pm 2.80$} \\
\midrule
Average            & \textbf{$28.33 \pm 2.37$}  & \textbf{$41.01 \pm 3.44$} \\
\bottomrule
\end{tabular}
\label{tab:temporal}
\end{table}

\begin{figure}[t]
    \centering
    \includegraphics[width=0.9\linewidth]{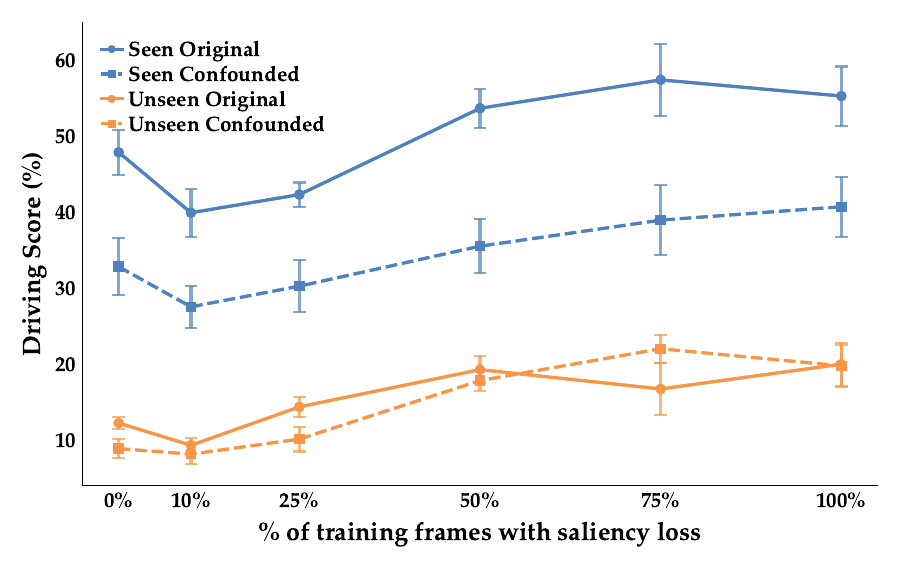}
    \caption{\textbf{Saliency fraction sweep.}
    Performance across Seen/Unseen and Original/Confounded splits.}
    \label{fig:fraction_curve}
    \vspace{-5mm}
\end{figure}

\textbf{Fraction of frames with saliency supervision.} In \autoref{fig:fraction_curve}, we vary the fraction $f\in\{0\%,10\%,25\%,50\%,75\%,100\%\}$ of frames that receive saliency regularization, while all frames are always trained with the BC loss. Performance increases with higher fractions of supervised frames, saturating around 75\%, suggesting that moderate coverage of saliency supervision may be sufficient to guide attention effectively.

\section{Real-Robot Experiments}
\label{sec:real_robot}

\subsection{Platform and Tasks}
\begin{figure*}[t]
    \centering
    \includegraphics[width=0.95\linewidth]{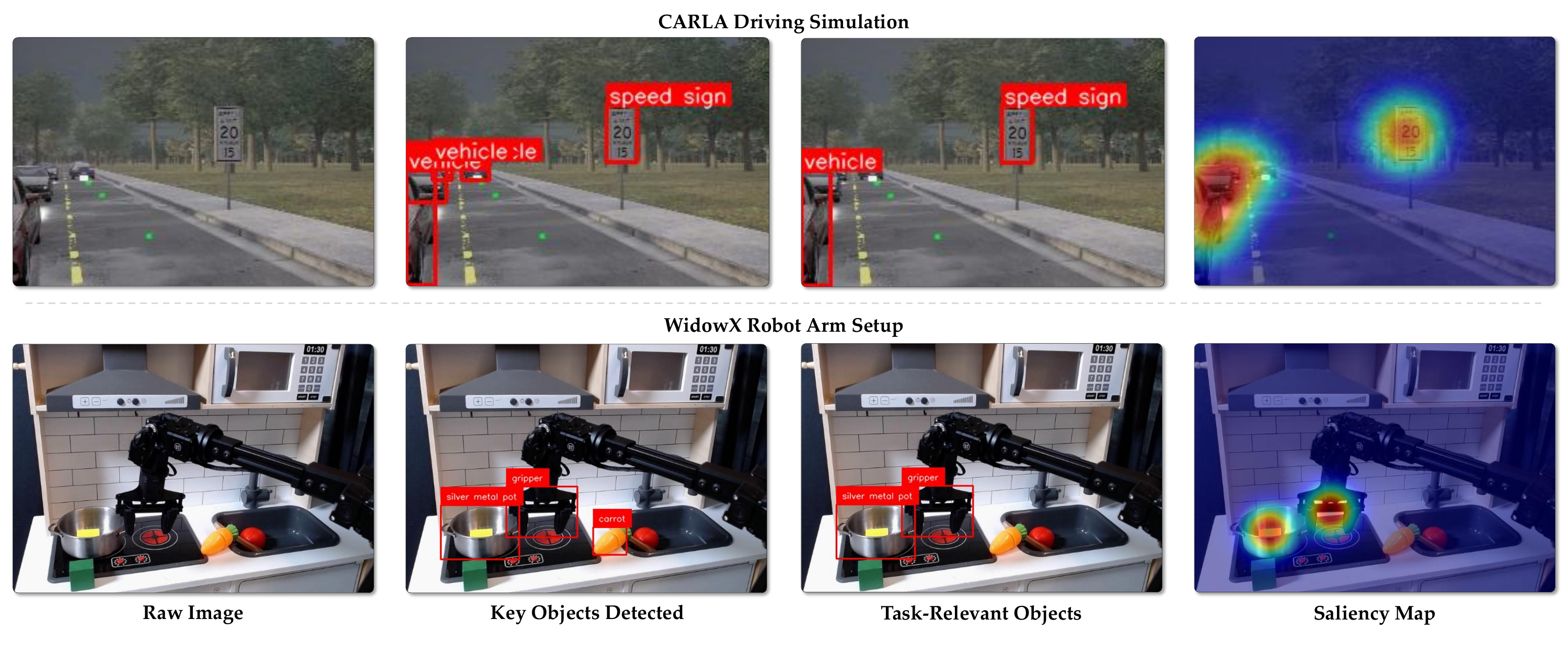}
    \vspace{-15px}
    \caption{Visualizations of our experimental setups and data processing steps. For example, in the WidowX Robotic Arm setup, the task is to pull a pot over the oven. Although the scene contains many distractors, AutoFocus-IL attends only to the most task-relevant objects.}
    \label{fig:demos}
\end{figure*}

We test AutoFocus-IL on a \textbf{WidowX} robot arm with a single monocular RGB camera at a resolution of $640\times480$ (see \autoref{fig:demos} for a visualization). Training images are center-cropped and resized to \(256{\times}256\). We evaluate on two manipulation tasks: \emph{Lift Carrot} and \emph{Pull Pot}. For each task, we also design a confounded variant by placing task-irrelevant distractor objects in the background to induce visual causal confusion. We consider the setup as four distinct single tasks, each trained with $100$ demonstrations.

\subsection{Policy and Saliency Integration}
At each timestep, the policy receives the latest two frames as input. All methods share the same vision backbone and optimization setup for a fair comparison.
The policy uses a {ResNet-34} \cite{he2016residual}  image encoder whose last convolutional feature map is flattened and concatenated with the 7D end-effector state, then passed to a Gaussian MLP head to parameterize a Gaussian action distribution. The saliency regularization weight is \(\lambda=5\). We let $\alpha=0.7$, $\beta=0.8$, $\gamma=30$, $t'=4$.

We additionally fine-tuned the YOLO \cite{redmon2016you} object detector to specifically detect the gripper of the robot arm, making up for the fact that Grounding DINO fails to detect the gripper.  

\subsection{Evaluation Protocol}
For each of the four settings, we perform 10 rollouts and report the number of successful episodes (task completion).
We compare AutoFocus-IL against Behavior Cloning (BC).

\subsection{Results and Discussion}
Results are summarized in \autoref{tab:real_robot_success}.
{AutoFocus-IL} consistently improves success on both tasks, with the largest gains under the confounded setting.

These trends mirror our simulation findings: VLM-driven, object-centric saliency helps the policy focus on causal scene elements and markedly improves robustness in the presence of unrelated visual clutter—without any additional human attention labels. Note that prior baselines such as GABRIL are not applicable to this task, as they require human gaze data—necessitating either specialized equipment for gaze collection or complex projection methods to map gaze onto the robot’s point-of-view camera.

\begin{table}[tb]
\centering
\caption{Real-robot manipulation task success rate}
\small
\renewcommand{\arraystretch}{1.2}
\begin{tabular}{lcc}
\toprule
\textbf{Tasks} & \textbf{AutoFocus-IL} & \textbf{BC} \\
\midrule
Lift Carrot & 7/10 & 5/10 \\
Lift Carrot Confounded & 7/10 & 1/10 \\
Pull Pot & 8/10 & 2/10 \\
Pull Pot Confounded & 6/10 & 0/10 \\
\midrule
\end{tabular}
\label{tab:real_robot_success}
\vspace{-3mm}
\end{table}

\section{Conclusion}
\label{sec:conclusion}

    \textbf{Summary.} In this study, we proposed a novel framework, AutoFocus-IL, which generates temporal saliency maps based on a vision-language model with zero human effort and uses them to regularize visual imitation learning. This framework significantly improves data efficiency and generalization capabilities, and can mitigate causal confusion in imitation learning. On the CARLA driving and real-world robotics tasks, it achieves stable improvements in both the \textit{Original} and \textit{Confounded} settings, and outperforms prior baselines that require human gaze, demonstrating its scalability and practicality.

    \noindent\textbf{Limitations and Future Work.} Our regularized training only considers CNN layers as a feature extractor. However, in recent years, the vision transformer (ViT) \cite{dosovitskiy2021image} backbone has been widely used in visual imitation learning \cite{zitkovich2023rt2,black2024$p_0$,shukor2025smolvla}. Recent research in computer vision has shown that regularization can enhance the performance of ViT models by improving the attention mechanism \cite{yang2025more,gandhi2025crisp}. Therefore, future work could focus on using our AutoFocus-IL framework to generate saliency maps for improving the performance of ViT-based policy models through regularized training. 
    
    Another limitation is that our method assumes that the VLM judgment from the first frame of each sub-sequence can represent object importance throughout the sub-sequence. However, in the real world, object states may change (e.g., a pedestrian entering the crosswalk), which may not be captured if detected objects remain the same, potentially causing the model to miss critical cues. This limitation also reflects the current VLM’s restricted ability to perceive motion and dynamic state transitions. Future work should explore richer contextual inputs \cite{mei2025survey} and more dynamic-aware VLMs \cite{azzolini2025cosmos} to better capture complex physical interactions.









\printbibliography

\end{document}